\documentclass[amsmath, amssymb, showkeys, titlepage, superscriptaddress, preprint, nofootinbib]{revtex4-2}

\usepackage{graphicx}
\usepackage{makecell}
\usepackage{booktabs}
\usepackage{algorithm}
\usepackage{amsmath,amssymb}
\usepackage{dsfont}
\usepackage{algpseudocode}
\usepackage{xcolor}
\usepackage[hidelinks]{hyperref}
\algrenewcommand\algorithmicindent{1.0em}
\algrenewcommand\algorithmicrequire{\textbf{Input:}}
\algrenewcommand\algorithmicensure{\textbf{Output:}}

\begin{document}

\title{COMO: Closed-Loop Optical Molecule Recognition with Minimum Risk Training}

\author{Zhuoqi Lyu}
\affiliation{Department of Data Science, College of Computing, City University of Hong Kong, Hong Kong, China}

\author{Qing Ke}
\email{q.ke@cityu.edu.hk}
\affiliation{Department of Data Science, College of Computing, City University of Hong Kong, Hong Kong, China}

\date{\today}

\begin{abstract}
Optical chemical structure recognition (OCSR) translates molecular images into machine-readable representations such as SMILES strings or molecular graphs, but remains challenging in real-world documents due to wide variations in chemical structures, shorthand conventions, and scanning artifacts. Most existing deep learning approaches rely on \emph{teacher forcing} with token-level Maximum Likelihood Estimation (MLE). This training paradigm suffers from exposure bias, as models are trained under ground-truth prefixes but must condition on their own previous predictions during inference. Moreover, token-level MLE objectives hinder the optimization towards molecular-level evaluation criteria such as chemical validity and structural similarity. Here we introduce Minimum Risk Training (MRT) to OCSR and propose COMO (\textbf{C}losed-loop \textbf{O}ptical \textbf{M}olecule rec\textbf{O}gnition), a framework that mitigates exposure bias by directly optimizing over molecule-level, non-differentiable objectives. Experiments on ten benchmarks, including synthetic and real-world chemical diagrams from patent and scientific literature, demonstrate that COMO substantially outperforms existing rule-based and learning-based methods with less training data, consistently achieving superior performance across both reproduced baselines and previously reported results. Ablation studies further show that MRT is architecture-agnostic, demonstrating its potential for broad application to end-to-end OCSR systems. 
\end{abstract}

\keywords{Optical chemical structure recognition, Molecule recognition, Minimum risk training, Non-differentiable optimization, Molecule modeling, Chemical informatics}

\maketitle

\section{Introduction}

Optical chemical structure recognition (OCSR) converts molecular structure images into machine-readable representations such as linear notations like SMILES strings~\citep{weininger1988smiles}. It is a fundamental methodology for mining large compound data sets, enabling the high-throughput digitization of molecules embedded within patents~\citep{morin2024patcid, papadatos2016surechembl} and scientific literature~\citep{rajan2023decimerai}. Unlocking this visually encoded chemical knowledge---which would otherwise remain inaccessible to computational analysis---is critical for propelling AI-driven drug discovery. Without reliable, automated interpretation of these visual disclosures, researchers face substantial bottlenecks in curating the vast datasets required to train generative models, develop foundation models, and advance multimodal neural networks in chemistry and the life sciences.

\begin{figure}[t]
\centering
\includegraphics[width=1\linewidth]{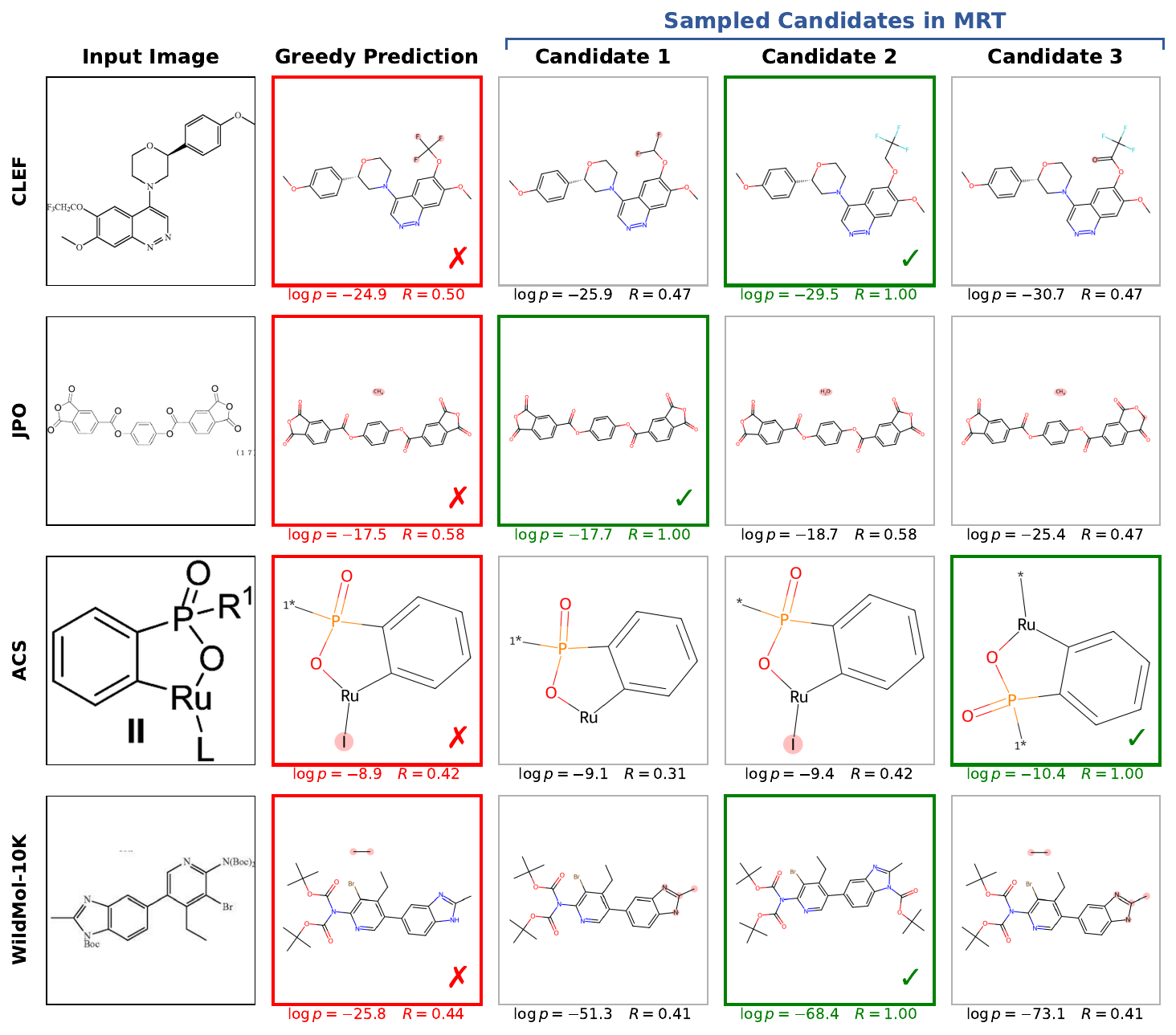}
\caption{\scriptsize Examples of predictions made by an MLE-trained model. Each row represents a sample from an OCSR dataset, followed by the corresponding model predictions. The first column shows the image inputs $x_i$. The second column presents greedy-search predictions $\hat{y_i}$, with the corresponding log probabilities $\log p(\hat{y_i} \mid x_i)$ (larger indicates higher probability) annotated below. In all shown examples, the greedy predictions fail (highlighted in red), with atom-level discrepancies explicitly highlighted in the skeletal formulas. In the third to the fifth columns, we display three prediction candidates generated using multinomial sampling ($N=32$, $\text{temperature}=0.5$). We show that the correct predictions (shown in green) can lie within the MLE model's search space. By applying meaningful similarity measurements (Morgan2-fingerprint-based Tanimoto similarity is used in this figure) between the candidate and the ground truth, we can compute a reward $R$ (annotated below) that serves as a guiding signal for backwardly optimizing the decoding distribution. When fed the same images into our trained COMO model, it always hits the correct predictions with the highest $R$. }
\label{fig:qualitative}
\end{figure}

Although recent deep-learning-based OCSR systems have substantially improved over rule-based pipelines by framing the task as image-to-sequence generation or graph prediction, accurate recognition from real-world documents remains challenging. As presented in Fig.~\ref{fig:qualitative}, models must cope with complex ring structures, substituent abbreviations (e.g., ``Et'', ``Me''), condensed formulas (e.g., ``F\textsubscript{3}CH\textsubscript{2}CO''), R-group placeholders (e.g., ``R\textsuperscript{1}'', ``L''), as well as cluttered surrounding text and visual artifacts introduced by scanning and document layout. These factors create a substantial distribution gap between the clean, synthetic images commonly used for training and the heterogeneous molecular depictions encountered in real documents, causing models to produce predictions that are structurally plausible but chemically incorrect.  

A central reason for this failure is the mismatch between how current encoder-decoder models are trained and how they perform inference. Most modern systems rely on an autoregressive sequence decoder trained with \emph{teacher forcing}, which minimizes token-level cross-entropy under ground-truth prefixes. While effective for learning local syntactic patterns, this Maximum Likelihood Estimation (MLE) objective introduces two compounding limitations: 
\begin{itemize}
\item \textbf{Exposure bias}: Under teacher forcing, the decoder is trained to condition on gold-standard prefixes, whereas at inference time, it must condition on its own previous outputs rather than ground-truth prefixes~\citep{bengio2015scheduled, ranzato2015sequence}. Consequently, errors made early in decoding can propagate and accumulate in ways never encountered during training. This issue is especially severe for SMILES generation: a single incorrect atom type, misplaced parentheses, or erroneous ring-closure digit can render an otherwise correct molecular structure chemically invalid, turning a near-miss into a complete failure. 
\item \textbf{Objective function mismatch}: The token-level cross-entropy objective is structurally incapable of optimizing the molecular-level metrics, including chemical validity, Tanimoto similarity, and exact structural match, which the evaluation actually measures. These metrics are inherently discrete and non-differentiable, and hence inaccessible to standard MLE training.
\end{itemize}

Fig.~\ref{fig:qualitative} illustrates these failures and points toward a remedy. Across several representative real-world examples, greedy decoding from a purely MLE-trained model yields incorrect predictions. However, when the same model is allowed to generate multiple candidates via sampling, the correct structure is often already present within its search space. The problem is therefore not a capacity failure: the model \emph{can} represent the correct answer. The problem is that MLE has concentrated probability mass on a suboptimal decoding path. If one could evaluate these candidates against a molecular-level similarity signal and redistribute probability mass toward the chemically superior ones, the training objective would directly align with the evaluation criterion---closing the loop between prediction quality and parameter update.

\begin{figure}[t]
\centering
\includegraphics[width=0.95\linewidth]{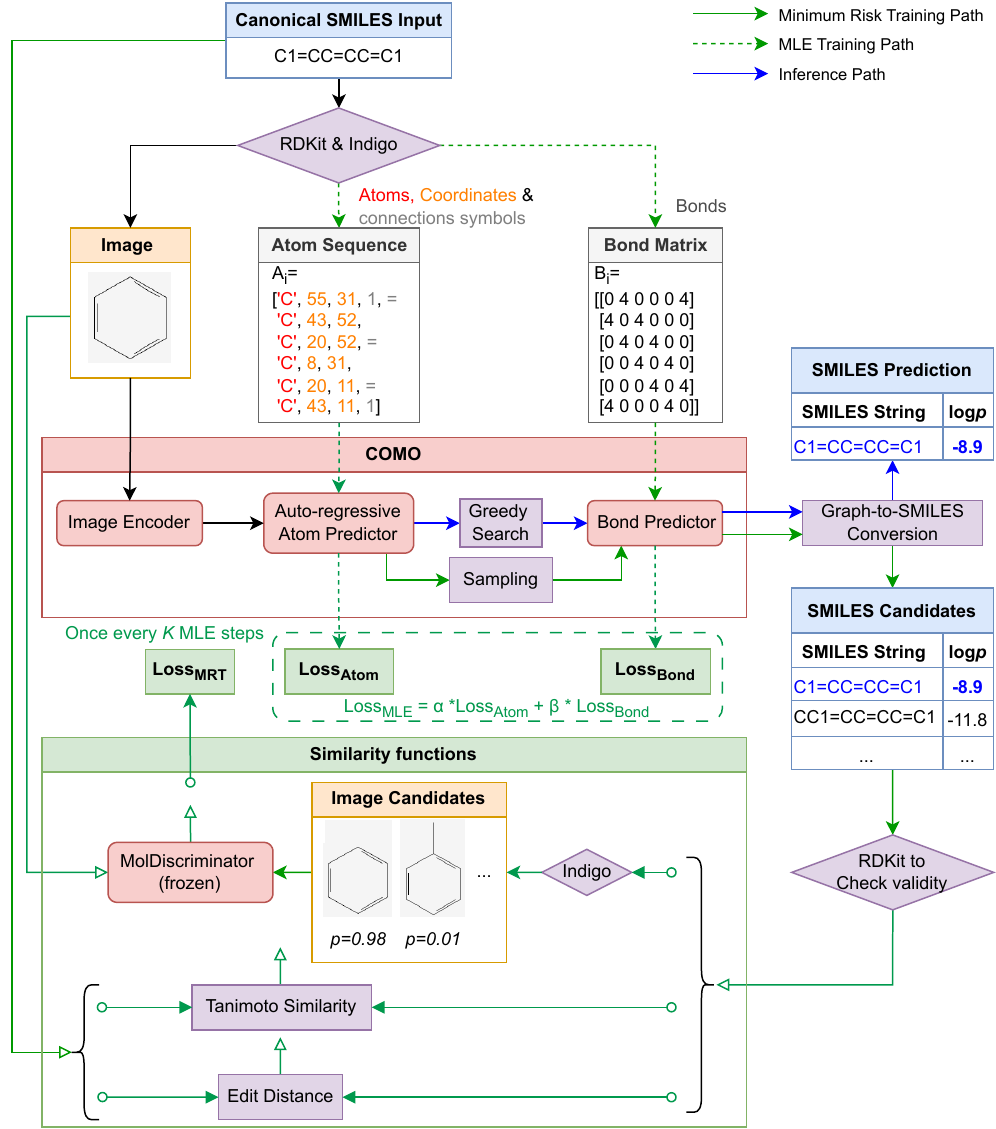}
\caption{Pipeline overview of COMO. Training paths are highlighted in green, and the inference path is highlighted in blue. Solid and dashed arrows indicate MRT and MLE, respectively. Hollow arrows indicate interchangeable similarity functions used to compute reward at the bottom. }
\label{fig:pipeline}
\end{figure}

Motivated by this observation, we propose \emph{COMO}, a \textbf{C}losed-loop \textbf{O}ptical \textbf{M}olecule rec\textbf{O}gnition framework built with \emph{Minimum Risk Training} (MRT)~\citep{shen2016minimum, och2003minimum}. Originally developed for neural machine translation and later adapted to image captioning and other sequence-generation tasks, MRT provides a principled way to directly optimize discrete, nondifferentiable metrics~\citep{rennie2017self}. We instantiate it here for molecular recognition. As shown in Fig.~\ref{fig:pipeline}, COMO retains the encoder-decoder backbone while augmenting purely token-level MLE with a joint MLE--MRT training strategy. During MRT, the model samples candidate SMILES from its own predictive distribution, evaluates them using a chemical-aware reward, and updates its parameters to favor candidates with higher reward. The reward function is modular, combining three complementary signals: chemical validity, exact string match, and an interchangeable structural similarity term, which can be instantiated as edit distance, fingerprint-based Tanimoto similarity, or a learned visual similarity score from a separate image encoder (MolDiscriminator). 

This design offers three practical advantages. First, by computing gradients over samples drawn from the model's own predictive distribution rather than teacher-forced prefixes, MRT directly closes the exposure-bias loop: the distribution the model optimizes during training matches the distribution it operates under at inference. Second, by defining the loss in terms of a molecular-level reward rather than token-level log-probabilities, MRT can optimize non-differentiable objectives such as chemical validity, structural similarity, and exact match directly, which is not achievable in MLE training. Third, reward-based supervision requires only ground-truth SMILES strings, without the need for atom coordinates or bond annotations, enabling COMO to benefit from the large pool of real-world molecular images that lack dense labels. 

Extensive experiments demonstrate the effectiveness of the proposed approach. COMO achieves state-of-the-art accuracy across all ten OCSR benchmarks, with particularly large improvements compared with prior-best baselines on challenging real-world datasets, including ACS (+15.1\,pp, to MolScribe), JPO (+10.2\,pp, to MolParser), CLEF (+4.2\,pp, to MolParser), and Staker (+4.1\,pp, to MolSight). Notably, these gains are obtained using substantially less training data than the strongest prior system for real-world settings, MolParser \citep{fang2025molparser}. Ablation studies further confirm that the gains stem from the training mechanism rather than from additional data. Moreover, applying the same MRT procedure to a structurally distinct model, SwinOCSR, yields accuracy gains of 2--20$\times$ on real-world benchmarks, confirming that MRT is a model-agnostic training principle for OCSR.

To the best of our knowledge, our work is the first effort to introduce MRT into end-to-end OCSR. In summary, our work makes three main contributions. 
\begin{itemize}
\item We propose COMO, a closed-loop OCSR framework that incorporates minimum risk training into an encoder--decoder pipeline, directly bridging the gap between token-level training and molecular-level evaluation objectives. 
\item We design a modular reward function with three variants: string-level edit distance, fingerprint-based Tanimoto similarity, and a visual similarity score produced by a self-supervised Siamese encoder (MolDiscriminator), enabling reward computation without requiring coordinate annotations. 
\item We demonstrate that MRT is a data-efficient, architecture-agnostic approach to real-world OCSR, outperforming prior methods across 10 benchmarks while using substantially fewer training samples.
\end{itemize}

We have made our trained models publicly available at \url{https://huggingface.co/Keylab/COMO}. To facilitate their use, we have also developed a Python package, \texttt{como-ocsr}, for chemical diagram recognition, which is available at \url{https://pypi.org/project/como-ocsr}.  

\section{Related Work}

\subsection{Optical Chemical Structure Recognition}

\subsubsection{Rule-based methods}

Early OCSR systems rely on hand-crafted pipelines: molecular images are binarized and vectorized, after which bonds and atoms are assembled using geometric heuristics. CLiDE~\citep{ibison1993chemical} represents one of the first automated efforts, followed by OSRA~\citep{filippov2009optical} and MOLVEC~\citep{peryea2019molvec}, which together became the mainstream automatic solutions for over a decade. While competitive on clean synthetic images, these rule-based methods are brittle, struggling with scanning artifacts, stylistic variation, abbreviations, and complex ring systems that deviate from their hard-coded assumptions; their accuracy drops sharply on real-world patent and literature images \citep{rajan2020review}.

\subsubsection{Deep-learning-based methods}

With the development of deep-learning-based techniques, OCSR is reframed as an image-to-sequence task that directly generates linear notations (e.g., SMILES), typically via an encoder--decoder architecture. Img2Mol~\citep{clevert2021img2mol} encodes images into a fixed latent vector and decodes SMILES from it; however, it shows low accuracy on structurally complex molecules. SwinOCSR~\citep{xu2022swinocsr} pairs a Swin Transformer encoder with a Transformer decoder but relies on synthetic training data only, hindering its real-world generalization. DECIMER~\citep{rajan2023decimerai} evolves from an ``ImageNet-RNN'' to an ``EfficientNet-Transformer'' architecture over iterations and extends to an end-to-end system to detect and recognize chemical structures.

In parallel, graph-based approaches are proposed to predict explicit molecular graph structures. ChemGrapher~\citep{oldenhof2020chemgrapher} pioneers this formulation using a cascade of segmentation and classification networks. MolMiner~\citep{xu2022molminer} detects atoms and bonds as objects using a YOLO-style detector and assembles the resulting graph by distance rules. MolGrapher~\citep{morin2023molgrapher} decomposes recognition into keypoint detection, graph construction, and GNN classification, achieving strong results on classic benchmarks. MolScribe~\citep{qian2023molscribe} formulates OCSR as image-to-graph prediction, using a Pix2Seq-style~\citep{chen2022pix2seq}  decoder to jointly predict atom tokens and 2D coordinates and a feedforward network for bond types, then constructing molecular graphs from the predicted atoms and bonds, where chemical constraints can be applied. This method significantly outperforms prior approaches across multiple benchmarks. 

\subsubsection{Large-scale training and vision--language models approaches}

Recent work scales training data aggressively. MolParser~\citep{fang2025molparser} adopts a Swin--BART architecture and trains on 7.7 million samples---including 400K human-annotated real-world images---establishing a new state of the art on real-world benchmarks. MolSight~\citep{zhang2026molsight} employs an EfficientViT encoder, scales MolScribe-style training on a portion of MolParser-7M, and further introduces GRPO-based reward post-training on a stereoisomer-focused dataset: a group-relative policy gradient over $G{=}4$ sampled completions, optimized with a weighted combination of Tanimoto similarity and a hand-crafted, coarse-grained stereochemical score. Its released checkpoints, however, indicate that this stage has not yet translated into general gains: exact-match accuracy on USPTO declines from 91.2 to 90.3 after GRPO\footnote{\url{https://github.com/hustvl/MolSight/issues/1}}, and the best released model (92.0), which underlies its reported benchmark results, is trained without RL\footnote{\url{https://github.com/hustvl/MolSight}}. For comparison, in our proposed method, MRT is interleaved with MLE throughout the main training schedule, minimizing a self-normalized expected risk (Eq.~\ref{alg:mrt_loss}) over $N{=}32$ probability-sharpened candidates under a hierarchical composite reward, yielding consistent real-world gains. 

Concurrently, large pretrained vision-language models (VLMs) have been explored for OCSR. Fine-tuned on 1 million PubChem molecules, PaliGemma~2~\citep{steiner2024paligemma} achieves 94.8\% exact match on a synthetic benchmark, demonstrating that large-scale VLM pretraining can transfer to OCSR without a chemistry-specific architecture. Uni-SMART~\citep{cai2024uni} takes a broader approach, jointly training a multimodal scientific literature model on molecules, reactions, charts, and tables from scientific publications, but is not designed or evaluated as a dedicated OCSR system. ChemVLM~\citep{li2025chemvlm} combines a vision transformer with a domain-specific language model and is trained on a curated bilingual dataset of molecules, reactions, and exam problems; while it outperforms general-purpose VLMs on OCSR, its exact-match accuracy remains well below specialized encoder--decoder systems such as MolScribe. Similarly, MolSeek-OCR~\citep{tang2026fine} fine-tunes DeepSeek-OCR-2 on a MolScribe-alike corpus comprising both rendered and realistic molecular images from patents, yet yields inferior performance compared to dedicated models like MolScribe. 

Despite this architectural diversity, no prior OCSR systems have successfully optimized molecular-level metrics directly during training. Furthermore, sequence-based methods rely on teacher-forced token-level MLE, which additionally suffers from \emph{exposure bias}. Graph-based detection and classification models (ChemGrapher, MolMiner, MolGrapher) avoid sequential decoding but minimize per-atom and per-bond classification losses that are equally disconnected from molecular-level chemical correctness criteria. COMO addresses both limitations through introducing Minimum Risk Training. 

\subsection{Exposure Bias and Minimum Risk Training}

The mismatch between training under gold-standard prefixes and inference over the model's own predictions is known as \emph{exposure bias} in sequence generation~\citep{bengio2015scheduled, ranzato2015sequence}. This is addressed by scheduled sampling~\citep{bengio2015scheduled}, which gradually substitutes ground-truth tokens with model-generated tokens during training via a curriculum schedule. While it reduces the training--inference gap, it retains token-level cross-entropy loss and therefore cannot directly optimize sequence-level objectives. 

A parallel line of work applies policy gradient methods to sequence generation, enabling direct optimization of non-differentiable, sequence-level objectives. \citet{ranzato2015sequence} proposed MIXER, which uses REINFORCE to optimize BLEU or ROUGE in text generation. \citet{rennie2017self} introduced Self-Critical Sequence Training (SCST) for image captioning, using the model's own greedy-decoded output as a variance-reduction baseline for REINFORCE; this allows direct optimization of CIDEr and yields large improvements over MLE-trained captioning models.

Minimum Risk Training (MRT) provides a complementary perspective rooted in statistical decision theory. The foundational instantiation in natural language processing is Minimum Error Rate Training \citep{och2003minimum}, which minimizes expected word error over an $N$-best list in phrase-based machine translation. Extending this to neural machine translation, \citet{shen2016minimum} samples $N$ candidates from its own predictive distribution at each training step rather than a fixed $N$-best list, computes sharpened importance weights, and updates parameters to minimize expected risk. The key advantage of this formulation over single-sample REINFORCE is that averaging over $N$ weighted samples substantially reduces gradient variance, leading to more stable training. Similar formulations of reward-based sequence optimization have since been applied to abstractive summarization \citep{paulus2018deep} and speech recognition~\citep{prabhavalkar2018minimum}. 

\subsection{Challenges of MRT for OCSR}

To the best of our knowledge, no prior work has introduced minimum risk training into end-to-end OCSR. Although widely studied in machine translation, directly applying existing MRT formulations to OCSR is far from straightforward. OCSR poses several unique challenges that are fundamentally distinct from natural language generation. First, unlike natural language generation, where errors typically degrade outputs smoothly, small token-level deviations in SMILES produce syntactically invalid or chemically implausible molecules, resulting in collapsing molecular validity entirely and consequently extremely sparse learning signals. Second, exposure bias is more severe in OCSR, as early errors during decoding can cause locally plausible predictions to become globally incorrect, widening the gap between token-level training and molecular-level evaluation. Third, evaluation objectives are non‑differentiable and non‑decomposable. Molecular correctness can depend on graph structure, similarity, or visual consistency, none of which decompose naturally over tokens, making standard reinforcement learning or policy gradient methods suffer from high variance and unstable training when applied directly to OCSR models. 

In this work, COMO proposes a closed-loop MRT framework specifically adapted to these challenges. We design a molecular-level reward function for COMO (chemical validity, Tanimoto similarity, exact structural match), which are precisely the type of discrete, non-differentiable objectives that MRT is designed to optimize. By sampling complete molecular candidates, computing molecular-level rewards, and integrating MRT with token-level MLE, COMO aligns the learning objective with actual inference operations while maintaining training stability. Moreover, we demonstrate that this formulation is architecture-agnostic and can be applied to diverse OCSR backbones without modifying their decoding mechanisms. 

To summarize, COMO is the first system to explore the connection between MRT and OCSR, and it is not a direct, straightforward reuse of existing MRT techniques, but rather a principled adaptation that addresses challenges in sequence generation for OCSR.

\section{COMO}

\subsection{Architecture}

COMO employs an encoder–decoder architecture for image-to-graph molecular recognition~\citep{qian2023molscribe}. Given an input molecular image $I$, the model predicts a 2D molecular graph $G = (A, B)$, where $A = \{a_i = (l_i, x_i, y_i)\}$ denotes the set of atoms, each associated with a SMILES label $l_i$ and image-space coordinates $(x_i, y_i)$, and $B \subseteq A \times A \times \mathcal{T}$ denotes the set of typed bonds. The bond type set $\mathcal{T}$ includes single, double, triple, aromatic, and wedge bonds. 

\subsubsection{Image encoder}

An ImageNet-pretrained Swin-B backbone~\citep{liu2021swin} encodes the
$384{\times}384$ input into spatial feature maps, which are projected to
$d_\text{model}{=}256$ dimensions and augmented with 2D sinusoidal positional
encodings.

\subsubsection{Sequence decoder}

We employ a 6-layer Transformer decoder with 8 attention heads to autoregressively generate
a \emph{chartok\_coords} token sequence
$S^A = [l_1, \hat{x}_1, \hat{y}_1, \ldots, l_n, \hat{x}_n, \hat{y}_n]$,
where each atom's SMILES sub-string $l_i$ is followed by its discretized coordinates
$\hat{x}_i, \hat{y}_i \in \{0,\ldots,63\}$. Structural tokens such as parentheses
and ring-closure digits are preserved in the sequence to help the decoder model
molecular topology. During training, the decoder is optimized via teacher forcing
with label-smoothed cross-entropy ($\epsilon{=}0.1$).

\subsubsection{Bond predictor}

For each atom $a_i$, the hidden state $\mathbf{h}_{a_i}$ at its final coordinate
token in the decoder output serves as the atom representation. A two-layer MLP
with GELU activation takes the concatenated pairs $[\mathbf{h}_{a_i}; \mathbf{h}_{a_j}]$
to classify bond types $b_{i,j}$ among seven classes (no bond, single,
double, triple, aromatic, solid-wedge, and dashed-wedge). Bidirectional predictions $b_{i,j}$ and $b_{j,i}$ are
symmetrized at inference time to improve accuracy. Stereochemistry is determined
by applying RDKit rules to the predicted graph and coordinates. 

\subsubsection{MolDiscriminator}

In addition to the recognition model described above, we train a separate \emph{MolDiscriminator} that will later serve as a visual reward signal for MRT. MolDiscriminator is a Siamese image-embedding network with an EfficientNet-V2-S backbone~\citep{tan2021efficientnetv2} followed by a projection head (Linear $\to$ ReLU $\to$ Linear) that maps pooled 1280-D features to $\ell_2$-normalized 128-D embeddings. Both branches share all parameters. It is trained with the symmetric InfoNCE loss~\citep{chen2020SimpleFramework} (temperature $\tau{=}0.07$): for each molecule, two views are rendered with randomly chosen styles (e.g., Indigo or RDKit, varying line widths and fonts), plus hard negatives generated by CReM~\citep{polishchuk2020crem} (radius =3) to supply structurally similar but distinct molecules. We train the network on 500K PubChem SMILES for 3 epochs. The trained model is frozen and provides a cosine-similarity-based visual reward that compares rendered images of a predicted SMILES $\hat{y}$ and the ground-truth $y^*$.

\subsection{Minimum Risk Training} \label{subsec:mrt}

As mentioned above, under standard MLE training, the decoder is trained via teacher forcing, which introduces \emph{exposure bias}: at inference time, predictions are conditioned on the model's own outputs (potentially erroneous) rather than ground-truth prefixes, leading to accumulated errors. More importantly, the MLE objective operates at the token level and cannot directly optimize molecular-level metrics, such as Tanimoto similarity. It also cannot leverage real-world images that lack atom-coordinate annotations, since MLE requires fully tokenized supervision.

To address these limitations, we adapt MRT~\citep{shen2016minimum, och2003minimum}, which minimizes the expected cost under the model's own sampling distribution:
\begin{equation}
\label{eq:mrt}
\mathcal{L}_{\text{MRT}}(\theta) = \mathbb{E}_{\hat{y} \sim p_\theta(\cdot \mid x)}\bigl[\Delta(\hat{y},\, y^*)\bigr] \approx \sum_{i=1}^{N} {Q}_\alpha(\hat{y}_i \mid x) \cdot \bigl(1 - R(\hat{y}_i,\, y^*)\bigr),
\end{equation}
where $\hat{y}_i$ are $N$ candidate SMILES sampled by multinomial decoding, ${Q}_\alpha(\hat{y}_i \mid x) \propto \exp\!\bigl(\alpha \cdot \log p_\theta(\hat{y}_i \mid x)\bigr)$ is a distribution of probability-sharpened importance weights, and $R \in [0,1]$ is a composite reward. 

\begin{algorithm}[H]
\scriptsize
\caption{ComputeMRTLoss}\label{alg:mrt_loss}
\begin{algorithmic}[1]
\Require 
    Encoded features $\mathbf{h} \in \mathbb{R}^{B' \times d \times H' \times W'}$, 
    ground-truth SMILES $\{y^*_b\}_{b=1}^{B'}$,

    model $\theta{=}\{\theta_{\text{enc}}, \theta_{\text{dec}}, \theta_{\text{bond}}\}$, frozen MolDiscriminator $g$, 
    
    sample size $N$, temperature $\tau$, sharpness $\alpha$, reward weights $w_v$, $w_s$, $w_e$
\Ensure MRT loss $\mathcal{L}_{\text{MRT}}$


\Statex \textit{// Phase 1: Sample candidates}
\State $\{\hat{y}_{b,i}\}_{i=1}^{N} \gets \textsc{SampleDecode}(\mathbf{h}_b, \tau;\; \theta_{\text{dec}})$ for each $b$ \Comment{multinomial sampling, no grad}


\Statex \textit{// Phase 2: Score candidates as inference-time}
\State $\text{logits}, \mathbf{u}_{b,i} \gets \textsc{Decoder}(\hat{y}_{b,i}, \mathbf{h}_b;\; \theta_{\text{dec}})$ \Comment{teacher-forced pass: logits for Phase 3, hidden states $\mathbf{u}_{b,i}$ for bond prediction}
\State $B_i \gets \textsc{BondPredict}(\mathbf{u}_{b,i};\; \theta_{\text{bond}})$ \Comment{$\mathbf{u}_{b,i}$ detached, no grad}
\State $m_i \gets \textsc{Postprocess}(\hat{y}_{b,i}, B_i)$ \Comment{graph-to-SMILES, abbreviation expansion}
\State $\text{valid}_i, \text{exact}_i \gets \textsc{Canonicalize}(m_i, y^*_b)$ \Comment{RDKit $\rightarrow$ validity \& exact match}
\State $s_i \gets \textsc{Sim}(m_i,\; y^*_b) \cdot \mathds{1}[\text{valid}_i]$ \Comment{See Eq.~\ref{eq:sim} for $\textsc{Sim}$ variants}
\State $R_i \gets w_v \cdot \mathds{1}[\text{valid}_i] + w_s \cdot s_i + w_e \cdot \mathds{1}[\text{exact}_i]$


\Statex \textit{// Phase 3: Teacher-forced log-probabilities}
\State $\log p(\hat{y}_i \mid \mathbf{h}_b) \gets \sum_{t=1}^{T_i} \log p(\hat{y}_{i,t} \mid \hat{y}_{<t}, \mathbf{h}_b)$
\Comment{with grad: $\theta_{\text{dec}}$, and $\theta_{\text{enc}}$}


\Statex \textit{// Phase 4: MRT loss}
\State ${Q}_{\alpha}(\hat{y}_i \mid \mathbf{h}_b) \gets \exp(\alpha \cdot \log p(\hat{y}_i \mid \mathbf{h}_b)) \big/ \sum_{j=1}^{N} \exp(\alpha \cdot \log p(\hat{y}_j \mid \mathbf{h}_b))$
\State $\mathcal{L}_{\text{MRT}} \gets \frac{1}{B'} \sum_{b=1}^{B'} \sum_{i=1}^{N} {Q}_\alpha(\hat{y}_i \mid \mathbf{h}_b) \cdot (1 - R_i)$
\end{algorithmic}
\end{algorithm}

As detailed in Algorithm~\ref{alg:mrt_loss}, the $\mathcal{L}_{\text{MRT}}$ is computed through the full inference-time pipeline: a pool of atom/coordinate sequence candidates is first constructed via multinomial decoding, then bond types are predicted greedily from the candidate's hidden states and the same postprocessing as at inference (graph-to-SMILES, abbreviation expansion, canonicalization) converts each candidate into its final molecule before scoring. Each candidate is therefore charged exactly the reward it would obtain at inference under the current model, and the MRT gradient updates the encoder and the atom/coordinate decoder; the bond predictor $\theta_{\text{bond}}$ receives no MRT gradient and is trained exclusively by the interleaved MLE steps (\S~\ref{sec:schedule}).

\subsubsection{Reward design}

We define a composite reward that contains three terms that capture progressively finer aspects of prediction quality:
\begin{equation}
\label{eq:reward}
R(\hat{y}, y^*) = w_v \cdot \mathds{1}[\text{valid}] + w_s \cdot \textsc{Sim}(\hat{y}, y^*) \cdot \mathds{1}[\text{valid}] + w_e \cdot \mathds{1}[\text{exact}],
\end{equation}
where $\mathds{1}[\text{valid}]$ indicates whether $\hat{y}$ is a chemically valid SMILES as verified by RDKit, $\mathds{1}[\text{exact}]$ checks canonical string equivalence, and $\textsc{Sim}$ is an interchangeable similarity function detailed in Eq.~\ref{eq:sim}.

The similarity component $\textsc{Sim}$ is designed to be modular: any differentiable or non-differentiable metric that scores a predicted SMILES against a reference can be plugged in.  We instantiate three variants that capture complementary aspects of molecular equivalence from a multi-modal perspective. 
\begin{equation}
\label{eq:sim}
\textsc{Sim}(\hat{y}, y^*) =
\begin{cases}
1 - \frac{\textsc{Lev}(\hat{y}, y^*)}{\max(|\hat{y}|, |y^*|)} & \text{(edit distance)} \\
\textsc{Tanimoto}\bigl(\textsc{FP}(\hat{y}), \textsc{FP}(y^*)\bigr) & \text{(Tanimoto)} \\
\cos\bigl(g(\textsc{Render}(\hat{y})), g(\textsc{Render}(y^*))\bigr) & \text{(visual)} \\
\end{cases}
\end{equation}
Levenshtein edit distance $\textsc{Lev}(\cdot,\cdot)$ treats SMILES as a string, focusing on sequential and syntactic similarity. Tanimoto similarity operates on molecular Morgan fingerprints $\textsc{FP}(\cdot)$, comparing chemical substructures represented as graphs. Here we use Morgan circular fingerprints~\citep{rogers2010extended}. The visual similarity leverages rendered depictions $\textsc{Render}(\cdot)$ generated from SMILES strings by Indigo or RDKit and evaluates molecules at the image (perceptual) level using embeddings $g(\cdot)$ encoded by the frozen image encoder MolDiscriminator\footnote{We frame visual reward primarily as a demonstration that the reward interface accepts learned perceptual signals rather than as a stronger reward; given its deliberately lightweight encoder, we interpret its performance under a representation-power caveat (\S~7).}. These formulations provide a diverse set of supervisory signals that span textual, structural, and perceptual representations, enabling MRT to assess molecular predictions across different modes of equivalence. We train one COMO model per similarity variant: the architecture, training data, and all hyperparameters are kept identical, and only the \textsc{Sim} term of the reward differs. We refer to the three resulting models as COMO-Edit, COMO-Tanimoto, and COMO-Visual, respectively. 

\subsubsection{Training schedule}
\label{sec:schedule}


As suggested in~\cite{shen2016minimum}, we warm up the model with several epochs of pure MLE before introducing MRT, since MRT relies on the model's own samples to compute rewards---starting from random weights would yield only invalid SMILES and zero reward. We further find that interleaving MRT with MLE (rather than switching to MRT exclusively after warmup) is essential: MLE retains the bond-prediction signal absent from MRT samples and stabilizes the encoder against distributional drift (empirically verified in \S~\ref{sec:twostage}, Table~\ref{tab:twostage}).

\begin{algorithm}[H]
\scriptsize
\caption{Training Pipeline}\label{alg:training_pipeline}
\begin{algorithmic}[1]
\Require
    MLE dataset $\mathcal{D}_{\text{MLE}}$, 
    MRT dataset $\mathcal{D}_{\text{MRT}}$, 
    model $\theta{=}\{\theta_{\text{enc}}, \theta_{\text{dec}}, \theta_{\text{bond}}\}$, 
    
    token weight $\alpha$, 
    bond weight $\beta$, 
    warm-up epochs $E_w$, 
    MRT weight $\lambda$, 
    
    MLE batch size $B$, 
    MRT batch size $B'$, 
    frozen MolDiscriminator $g$
\Ensure Trained $\theta$


\State $T \gets \lceil|\mathcal{D}_{\text{MLE}}| / B\rceil$
    \Comment{MLE steps per epoch}
\State $T_{\text{MRT}} \gets \lceil|\mathcal{D}_{\text{MRT}}| / B'\rceil$
    \Comment{MRT batches per epoch}
\State $K \gets \max\!\bigl(1,\,\lfloor T \,/\, T_{\text{MRT}}\rfloor\bigr)$
    \Comment{MRT frequency: one MRT step every $K$ MLE steps, so $\mathcal{D}_{\text{MRT}}$ is traversed ${\sim}$once per epoch}


\For{$e = 1, \dots, E$} \Comment{epoch loop}

\For{$t = 1, \dots, T$} \Comment{MLE step loop}
    \State Sample MLE batch $\{(I_j, s_j, \mathbf{c}_j, E_j)\}_{j=1}^{B}$ from $\mathcal{D}_{\text{MLE}}$
        \Comment{rendered image, SMILES, coords, bonds}
    \State $\mathbf{t}_j \gets \textsc{Tokenize}(s_j, \mathbf{c}_j)$ for each $j$
    \Statex \hspace{\algorithmicindent}\hspace{\algorithmicindent}\textit{// MLE forward + backward}
    \State $\hat{\mathbf{p}}_{\text{tok}},\, \hat{\mathbf{p}}_{\text{edge}},\, \_ \gets
        \textsc{Model}_\theta\bigl(\{I_j\},\, \{\mathbf{t}_{j,1},\, \mathbf{t}_{j,2},\, \dots,\, \mathbf{t}_{j,T-1}\}\bigr)$
        \Comment{Teacher-forcing}
    \State $\mathcal{L}_{\text{MLE}} \gets
        \alpha \cdot \textsc{CE}\bigl(\hat{\mathbf{p}}_{\text{tok}},\, \{\mathbf{t}_{j,2},\, \mathbf{t}_{j,3},\, \dots,\, \mathbf{t}_{j,T}\};\; \epsilon{=}0.1\bigr)
        + \beta \cdot \textsc{CE}\bigl(\hat{\mathbf{p}}_{\text{edge}},\, \{E_j\};\; \mathbf{w}_c\bigr)$
    \State $\nabla_\theta \mathcal{L}_{\text{MLE}}.\textsc{backward}()$
        \Comment{gradients accumulate in $\theta.\text{grad}$}
    \Statex \hspace{\algorithmicindent}\hspace{\algorithmicindent}\textit{// MRT forward + backward (interleaved every $K$ MLE steps, skip during warmup)}
    \If{$e > E_w$ \textbf{and} $t \bmod K = 0$}
        \State Sample $\{(x_b, y_b^*)\}_{b=1}^{B'}$ from $\mathcal{D}_{\text{MRT}}$
        \State $\mathbf{h} \gets \textsc{PosEnc2D}\bigl(\textsc{Enc}_\theta(x)\bigr)$
            \Comment{with grad, shared encoder}
        \State $\mathcal{L}_{\text{MRT}} \gets \textsc{ComputeMRTLoss}(\mathbf{h},\, \{y_b^*\},\, g;\; \theta)$
            \Comment{Alg.~\ref{alg:mrt_loss}}
        \State $\nabla_\theta (\lambda \cdot \mathcal{L}_{\text{MRT}}).\textsc{backward}()$
            \Comment{gradients accumulate with MLE grads}
    \EndIf
    \Statex \hspace{\algorithmicindent}\hspace{\algorithmicindent}\textit{// Optimizer step on accumulated gradients}
    \State $\theta \gets \theta - \textsc{AdamW}\bigl(\nabla_\theta;\;
        \text{clip}\;\lVert\nabla\rVert \le 1\bigr)$
\EndFor
\EndFor
\end{algorithmic}
\end{algorithm}

As detailed in Algorithm~\ref{alg:training_pipeline}, the first $E_w$ epochs perform MLE-only warmup to establish meaningful encoder representations; subsequent epochs insert one MRT step every $K$ MLE steps so that $\mathcal{D}_{\text{MRT}}$ is traversed approximately once per epoch. Gradients from both objectives accumulate before a single optimizer step, yielding a combined update $\nabla_\theta\bigl(\mathcal{L}_{\text{MLE}} + \lambda \cdot \mathcal{L}_{\text{MRT}}\bigr)$. This design allows the model to learn from both fully supervised synthetic data and weakly supervised real-world images throughout training, while the MLE term anchors the decoder and maintains bond-predictor gradients that MRT alone cannot provide.

MolDiscriminator is trained offline, once, and independently of the recognition model. It remains frozen during MRT and serves as an optional, drop-in component of the reward.

\subsection{Training Data}

For MLE training, we construct a dataset $\mathcal{D}_{\text{MLE}}$ consisting of 1M SMILES randomly chosen from PubChem as synthetic samples and 680K molecules from patent grants as patent samples, provided by \citet{qian2023molscribe}, but with the following modifications: 

\begin{itemize}
    \item We remap the wildcard placeholders (``*'') in patent data from the original Mol files to match their visual representations (like ``R'', ``X'', and ``Me'') on images. This is aimed at aligning the training object of synthetic data and the patent data. We finally successfully recovered 652 thousand samples and used them in the MLE training.
    \item We apply an adaptive interpolation algorithm selection mechanism to independently scale up/down to enhance image quality during training and inference.
    \item We keep the original aspect ratio by padding the input molecule images rather than distorting them.
    \item We add a random comment in Japanese/Chinese, and simulated scan shadows as image augmentation.
    \item We randomly replace wedge bonds with wavy bonds to augment the molecular structure.
\end{itemize}

For MRT, we construct a dataset $\mathcal{D}_{\text{MRT}}$ based on MolParser-SFT, which contains 91.2K real-world molecular images with E-SMILE annotations \citep{fang2025molparser}. We exclude samples with position variation indicator (\textlangle r\textrangle), frequency variation indicator (?), and abstract ring indicator (\textlangle c\textrangle), as they are out of the scope of this study. We then expand and map abbreviated functional groups within the substituent indicator (\textlangle a\textrangle) back into the SMILES backbone. We treat all orphaned wildcard placeholders or connection points (\textlangle dum\textrangle) as ``*'', and obtain about 83K samples for MRT training.

\section{Experiments}

In this section, we evaluate our models on a comprehensive set of mainstream OCSR benchmarks and compare them with recent methods. We further conduct ablation studies on training data and model architecture to demonstrate the effectiveness and generality of our proposed approach. 

\subsection{Baselines}

We compare COMO with seven representative OCSR methods.
\begin{itemize}
\item \textbf{OSRA}~\citep{filippov2009optical} is a rule-based system based on image binarization, vectorization, and heuristic atom-bond assembly without any learned components. 
\item \textbf{Img2Mol}~\citep{clevert2021img2mol} maps CNN-extracted features into the CDDD latent space, from which SMILES are decoded.
\item \textbf{SwinOCSR}~\citep{xu2022swinocsr} is an end-to-end encoder--decoder model that pairs a Swin Transformer encoder with a Transformer decoder to generate DeepSMILES. 
\item \textbf{MolGrapher}~\citep{morin2023molgrapher} adopts a multi-stage pipeline consisting of keypoint detection, candidate graph construction, and GNN-based node and edge classification. 
\item \textbf{MolScribe}~\citep{qian2023molscribe} uses a Swin-B encoder and a Transformer decoder to autoregressively generate molecular graphs with atom coordinates. 
\item \textbf{MolParser}~\citep{fang2025molparser} formulated OCSR as an image-captioning task, using a Swin encoder and a BART decoder to generate Extended SMILES (E-SMILES). 
\item \textbf{MolSight}~\citep{zhang2026molsight} builds on a MolScribe-like architecture with an EfficientViT encoder, adds extra pretraining on MolParser-7M \citep{fang2025molparser}, and detaches atom coordinate predictions. 
\end{itemize}

\subsection{Benchmarks}

We conduct experiments on widely used OCSR benchmarks, including CLEF, JPO, UOB, USPTO \citep{rajan2020review}, and Staker \citep{staker2019molecular}. Note that previous studies have identified label errors in several benchmarks  \citep{xiongAExtractorSystemAutomatic2024, fang2025molparser}, and we thus adopt manually-corrected versions of CLEF, JPO, and UOB datasets from \cite{xiongAExtractorSystemAutomatic2024}. Furthermore, we find a pervasive phenomenon of mislabeling molecules with their tautomers in UOB and therefore adopt \cite{xiongAExtractorSystemAutomatic2024}'s manner to report results after tautomer-standardization. We further include: (1) two synthetic benchmarks (Indigo and ChemDraw) and a real-world benchmark (ACS) introduced in \cite{qian2023molscribe}; (2) a large-scale real-world benchmark, USPTO-10K, contributed by \cite{morin2023molgrapher}; and (3) a large-scale real-world benchmark labeled with E-SMILES, WildMol-10K, proposed in \cite{fang2025molparser}. 

For the benchmark dataset WildMol-10K, we find that some abbreviated functional groups like ``Boc'' (\textit{tert}-Butyloxycarbonyl) are marked as wildcard placeholders in the SMILES backbone and annotated literally in the extension part. This representation hinders a fair evaluation of our models' capability, as the models can give concrete predictions as ``[C](=O)OC(C)(C)C'' but the label may require a placeholder ``*''. Therefore, to ensure fair evaluation, we systematically expand and map these substituents back into their corresponding backbone to reconstruct valid SMILES strings. We also find index conflicts between the substituent extension and the placeholder in some samples. In total, we exclude 111 samples from the benchmark due to expansion failures or index conflicts, and all remaining samples are validated using RDKit.

For all benchmarks, we follow the evaluation protocol in \cite{qian2023molscribe}. Specifically, both predicted and ground-truth SMILES are canonicalized, R-groups are replaced with wildcards (``*'' or ``[d*]'', where d is an integer), and tetrahedral chirality is preserved, while cis–trans isomerism is ignored because it is often absent in ground-truth. Finally, we compute and report the string exact match as the evaluation metric. 

\subsection{Experimental Setup}

We implement COMO in PyTorch and train it on 4$\times$NVIDIA L40 GPUs. We adopt the interleaved MLE-MRT joint pipeline to train COMO for 30 epochs. Following the modular reward in Eq.~\ref{eq:sim}, we train three independent COMO models---COMO-Edit, COMO-Tanimoto, and COMO-Visual---which differ only in the Sim term of the reward (edit distance, fingerprint Tanimoto, or visual similarity from MolDiscriminator). All remaining settings below are shared across the three runs. The MLE branch uses AdamW with a learning rate of $4{\times}10^{-4}$ for both the encoder and the decoder, weight decay of $10^{-6}$, a 2\% linear warmup followed by cosine decay, and a per-GPU batch size of 64. Input images are resized to $384{\times}384$ with aspect-ratio-preserving padding. The first 5 epochs are MLE-only warmup; after that, one MRT step is interleaved every $K{=}\lfloor T / T_{\text{MRT}} \rfloor$ MLE steps. The accumulated gradient norms are clipped to 1.0.

The MRT branch uses a per-GPU batch size of 16, $N{=}32$ candidate samples drawn by multinomial decoding at temperature $\tau{=}0.5$, sharpening exponent $\alpha{=}1.0$, which weights candidates by their model probabilities without additional distortion, maximum decoding length 500, and MRT loss weight $\lambda{=}0.1$, which keeps the combined update anchored to MLE. The reward weights assign a mild prerequisite role to validity ($w_v{=}0.1$), as sampled candidates are typically already valid; the dominant continuous signal to similarity ($w_s{=}0.5$), which is graded and most informative in the near-miss regime; and the largest per-occurrence bonus to exact match ($w_e{=}0.4$).

Regarding computational cost, per-epoch training time rises from about 1.5\,h under pure MLE to about 4.5\,h once MRT steps are interleaved---roughly a $3{\times}$ per-epoch cost during the 25 MRT-involved epochs. Inference cost is unchanged: at test time COMO performs a single greedy autoregressive decoding pass followed by bond prediction, exactly like an MLE-trained model, whereas alternatives such as beam search or ensembling multiply inference-time computation.

The visual reward encoder MolDiscriminator is trained separately on 500K PubChem SMILES for 3 epochs with a batch size of 48 per GPU, a learning rate of $3{\times}10^{-4}$, and a 3\% linear warmup.

For OSRA, Img2Mol, SwinOCSR, MolGrapher, MolScribe, and MolSight, we obtain their publicly released models for benchmarking. For MolSight, we use the MolSight-extra checkpoint, as it achieves the best reported performance among the three released models \citep{zhang2026molsight}. For MolParser, which is not publicly available, we use the benchmark results reported in the original paper \citep{fang2025molparser}. 

\subsection{Experimental Results}

\begin{table}
\centering
\scriptsize
\caption{Performance comparison with existing OCSR methods. Results are reported as exact-match accuracy, with the best performance in bold and the second-best in underlined.}
\label{tab:lead_board}
\begin{tabular}{{l}*{10}{c}}
\toprule
& \multicolumn{2}{c}{Synthetic} & \multicolumn{8}{c}{Real-World} \\
\cmidrule(lr){2-3} \cmidrule(lr){4-11}
& \makecell{Indigo\\(5719)} & \makecell{ChemDraw \\(5719)} & \makecell{CLEF\\(992)} & \makecell{JPO\\(450)} & \makecell{UOB\footnotemark[1]\\(5740)} & \makecell{USPTO\\(5719)} & \makecell{USPTO-10K\\(10000)} & \makecell{Staker\\(50000)} & \makecell{ACS\\(331)} & \makecell{WildMol-10K\\(10000)} \\
\midrule

OSRA 2.1\footnotemark[2] & $95.2$ & $84.7$ & $86.0$ & $58.6$ & $85.3$ & $88.1$ & $89.8$ & $0.0$ & $49.2$ & $24.9$\\

Img2Mol\footnotemark[2] & $59.1$ & $46.4$ & $17.7$ & $17.6$  & $84.6$ & $26.4$ & $24.9$ & $17.1$ & $23.6$ & $19.8$\\

SwinOCSR\footnotemark[2] & $67.8$ & $74.5$ & $26.9$ & $12.7$ & $ 49.3$ & $ 24.9$ & $ 32.4$ & $0.2$ & $20.2$ & $10.6$\\

MolGrapher\footnotemark[2] & 76.4 & 76.5 & 90.5 & 67.5 & 94.9 & 91.5 & 93.3 & 0.0 & 41.4 & 45.5 \\

MolScribe\footnotemark[2] & $97.4$ & $93.5$ & $84.6$ & $59.2$ & $94.8$ & $89.6$ & $91.0$ & $81.4$ & $73.1$ & $59.2$\\

MolSight\footnotemark[2] & 96.3 & 93.7 & 88.3 & 70.4 & 95.6 & 93.2 & 89.4 & 83.4 & 72.9 & 9.0 \\

MolParser-Tiny\footnotemark[3] & $-$ & $-$ & $91.0$ & $75.6$ & $91.6$ & $93.0$ & $89.5$ & $-$ & $-$ & $73.1$\\
MolParser-Small\footnotemark[3] & $-$ & $-$ & $90.8$ & $76.2$ & $91.1$ & $93.1$ & $94.8$ & $-$ & $-$ & $76.3$\\
MolParser-Base\footnotemark[3] & $-$ & $-$ & $90.7$ & $78.9$ & $91.8$ & $93.0$ & $94.5$ & $-$ & $-$ & $76.9$\\

\textbf{COMO-Edit} & $\mathbf{98.8}$ & $\underline{96.3}$ & $\mathbf{95.0}$ & $\mathbf{89.1}$ & $\mathbf{98.5}$ & $93.0$ & $95.9$ & $\mathbf{87.5}$ & $\mathbf{88.2}$ & $\mathbf{77.2}$\\
\textbf{COMO-Tanimoto} & $98.6$ & $\mathbf{96.5}$ & $\underline{94.8}$ & $\underline{88.4}$ & $\underline{98.0}$ & $\mathbf{93.4}$ & $\mathbf{96.1}$ & $\underline{87.4}$ & $84.6$ & $\underline{77.1}$\\
\textbf{COMO-Visual} & $\underline{98.7}$ & $\underline{96.3}$ & $94.6$ & $87.1$ & $96.8$ & $\underline{93.3}$ & $\underline{96.0}$ & $\mathbf{87.5}$ & $ \underline{86.4}$ & $76.9$\\
\bottomrule
\end{tabular}
\footnotetext[1]{Results on UOB are after tautomer standardization.}
\footnotetext[2]{results using released models.}
\footnotetext[3]{results from \cite{fang2025molparser}.}
\footnotetext{$-$: not available.}
\end{table}

Table~\ref{tab:lead_board} reports exact-match accuracy across 10 benchmarks. COMO results refer to the three reward-variant models of Eq.~\ref{eq:sim}. Overall, COMO achieves the best performance on every dataset when taking the strongest of the three reward variants, and the three variants deliver consistently competitive results with each other (spreads within 2 pp on nine of ten benchmarks; 3.6 pp on ACS), and we do not interpret small per-benchmark differences as a ranking of reward choices; the consistency across the three variants suggests that MRT is robust to the choice of similarity measurement. On the two synthetic sets, COMO is on par with or slightly above the previous best (MolScribe) (roughly 1--3 points); given run-to-run variability, we do not claim an improvement on this margin. On the real-world benchmarks, the improvement is particularly striking on JPO ($89.1\%$ vs.\ $78.9\%$, the previous best by MolParser-Base) and ACS ($88.2\%$ vs.\ $73.1\%$ by MolSight).

COMO improves over MolScribe on every benchmark: with macro-average, MolScribe reaches 82.4\% exact match, whereas COMO achieves 91.4\%--92.0\%, representing a consistent +9.0 to +9.6 point gain in overall accuracy. The gains range from modest on synthetic data (+1.4 on Indigo, +3.0 on ChemDraw) to substantial on real-world data (+10.4 on CLEF, +29.9 on JPO, +3.7 on UOB, +3.8 on USPTO, +5.1 on USPTO-10K, +6.1 on Staker,  +15.1 on ACS, and +18.0 on WildMol-10K). This larger gain on real-world benchmarks indicates that MRT is especially effective at bridging the domain gap between synthetic training images and real-world test images, which cannot be achieved by MLE alone, even when given the same additional real-world data (cf.\ ablation in Table~\ref{tab:ablation}).

MolParser is the strongest prior method on real-world benchmarks, benefiting from a 7.7-million-sample training set that includes 400K human-annotated real-world images curated through an active-learning data engine. In contrast, COMO uses only 1.65M MLE samples (1M PubChem + 652K USPTO) and 83K real-world images for MRT, roughly 4.7$\times$ less data in total. Despite this data disadvantage, COMO-Tanimoto outperforms MolParser-Base on CLEF, JPO, UOB, USPTO, USPTO-10K, and WildMol-10K. Note that MolParser is not open-source at the time of writing, and hence we cannot reproduce its results on benchmarks where the authors did not report numbers (Indigo, ChemDraw, Staker, ACS). Moreover, MolParser's evaluation protocol for WildMol-10K---in particular, how E-SMILES extensions are mapped back to canonical SMILES for comparison---is not publicly documented, making a fully apples-to-apples comparison on that benchmark difficult.

Several earlier methods exhibit sharp drops on specific datasets. OSRA performs well on clean synthetic images (95.2 on Indigo) but collapses on Staker (0.0) and WildMol-10K (24.9), highlighting brittleness under distribution shift. Img2Mol achieves consistently low accuracy, potentially due to the information bottleneck imposed by its 512-dimensional CDDD latent space. SwinOCSR, despite using a Swin-based encoder, generalizes poorly to real-world data, as it is trained exclusively on synthetic images. MolGrapher performs competitively on benchmarks like CLEF and USPTO-10K but struggles on Staker (0.0), ACS (41.4), and WildMol-10K (45.5), likely due to scale mismatches caused by small image sizes. Finally, MolSight suffers a drastic collapse on WildMol-10K despite large-scale pretraining, probably due to the prevalence of substituent and the absence of coordinate prediction. 

\begin{table}[h]
\centering
\scriptsize
\caption{Mean Tanimoto similarity (\%) between predictions and ground-truth molecules on the ten benchmarks. Same evaluation protocol as Table~\ref{tab:lead_board}.}
\label{tab:tanimoto}
\begin{tabular}{{l}*{10}{c}}
\toprule
& \multicolumn{2}{c}{Synthetic} & \multicolumn{8}{c}{Real-World} \\
\cmidrule(lr){2-3} \cmidrule(lr){4-11}
& \makecell{Indigo} & \makecell{ChemDraw} & \makecell{CLEF} & \makecell{JPO} & \makecell{UOB} & \makecell{USPTO} & \makecell{USPTO-10K} & \makecell{Staker} & \makecell{ACS} & \makecell{WildMol-10K} \\
\midrule
OSRA 2.1& $97.8$ & $90.5$ & $94.3$ & $70.0$ & $89.0$ & $94.8$ & $94.5$ & $4.3$ & $63.7$ & $45.9$\\
Img2Mol & $88.4$ & $80.4$ & $68.6$ & $43.4$ & $84.6$ & $65.3$ & $64.5$ & $55.7$ & $54.1$ & $56.8$\\
SwinOCSR & $87.9$ & $89.6$ & $56.1$ & $44.4$ & $87.1$ & $60.8$ & $65.6$ & $29.2$ & $40.1$ & $34.8$\\
MolGrapher & $97.6$ & $97.6$ & $89.2$ & $78.6$ & $92.6$ & $93.3$ & $95.9$ & $7.3$ & $67.3$ & $67.9$\\
MolScribe & $99.1$ & $96.9$ & $97.0$ & $86.2$ & $94.9$ & $98.2$ & $\underline{98.5}$ & $98.4$ & $89.2$ & $85.2$\\
MolSight & $99.1$ & $\underline{98.6}$ & $97.8$ & $88.3$ & $94.8$ & $\underline{98.5}$ & $97.3$ & $96.9$ & $89.9$ & $29.8$\\
\textbf{COMO-Edit} & $\textbf{99.6}$ & $\underline{98.6}$ & $\textbf{98.5}$ & $\textbf{95.0}$ & $\textbf{95.3}$ & $\underline{98.5}$ & $\textbf{98.8}$ & $\underline{98.5}$ & $\textbf{95.3}$ & $\textbf{94.7}$\\
\textbf{COMO-Tanimoto} & $\underline{99.5}$ & $98.5$ & $\underline{98.4}$ & $\underline{94.9}$ & $\textbf{95.3}$ & $\underline{98.5}$ & $\textbf{98.8}$ & $\textbf{98.6}$ & $\underline{94.3}$ & $\underline{94.4}$\\
\textbf{COMO-Visual} & $\underline{99.5}$ & $\textbf{98.7}$ & $98.2$ & $94.1$ & $\underline{95.2}$ & $\textbf{98.6}$ & $\textbf{98.8}$ & $\textbf{98.6}$ & $\textbf{95.3}$ & $\textbf{94.7}$\\
\bottomrule
\end{tabular}
\end{table}

As a secondary metric beyond exact match, Table~\ref{tab:tanimoto} reports the mean Tanimoto similarity between predictions and ground-truth molecules, computed under the identical evaluation protocol as Table~\ref{tab:lead_board}. The ranking mirrors that of Table~\ref{tab:lead_board}, and the near-zero Tanimoto scores of rule-based methods on degraded real-world data (e.g., Staker) reflect entirely invalid outputs rather than structurally close near-misses.

\subsection{Ablation Study} \label{sec:ablation}

\subsubsection{Effect of MRT}

To demonstrate that the observed performance improvement arises from the MRT training mechanism rather than from the additional 83K real-world images, we conduct an ablation study using the same data for pure MLE training. Specifically, we convert the E-SMILES strings in $\mathcal{D}_{\text{MRT}}$ back to standard SMILES, render them using Indigo to obtain images and coordinates, and train the model with MLE under the same interleaving schedule. All hyperparameters, including batch size, learning rate, number of epochs, number of warm-up epochs, and loss weights, are kept identical to those used in MRT (Algorithm~\ref{alg:training_pipeline}) to ensure a fair comparison. 

\begin{table}[h]
\centering
\scriptsize
\caption{Ablation study on the additional training data $\mathcal{D}_{\text{MRT}}$. Downward arrows indicate performance degradation when including $\mathcal{D}_{\text{MRT}}$ under pure MLE training, relative to training on $\mathcal{D}_{\text{MLE}}$ alone. The lower half presents results from MLE-MRT joint training using $\mathcal{D}_{\text{MLE}}+\mathcal{D}_{\text{MRT}}$, identical to those in Table~\ref{tab:lead_board}.}
\label{tab:ablation}
\begin{tabular}{{l}*{10}{c}}
\toprule
& \multicolumn{2}{c}{Synthetic} & \multicolumn{8}{c}{Real-World} \\
\cmidrule(lr){2-3} \cmidrule(lr){4-11}
& \makecell{Indigo} & \makecell{ChemDraw} & \makecell{CLEF} & \makecell{JPO} & \makecell{UOB} & \makecell{USPTO} & \makecell{USPTO-10K} & \makecell{Staker} & \makecell{ACS} & \makecell{WildMol-10K} \\
\midrule

\textit{Pure MLE} & & & & & & & & & & \\

Sole $\mathcal{D}_{\text{MLE}}$ & $98.5$ & $95.7$ & $93.0$ & $71.0$ & $96.5$ & $92.7$ & $96.2$ & $87.5$ & $75.8$ & $66.6$ \\
$\mathcal{D}_{\text{MLE}}+\mathcal{D}_{\text{MRT}}$ & $ 98.5$ & $94.8\downarrow$ & $92.4\downarrow$ & $72.6$ & $95.8\downarrow$ & $93.0$ & $96.1\downarrow$ & $87.4\downarrow$ & $77.6$ & $66.6$ \\

\midrule

\textit{MLE + MRT} & & & & & & & & & & \\
COMO-Edit & $\mathbf{98.8}$ & $\underline{96.3}$ & $\mathbf{95.0}$ & $\mathbf{89.1}$ & $\mathbf{98.5}$ & $93.0$ & $95.9$ & $\mathbf{87.5}$ & $\mathbf{88.2}$ & $\mathbf{77.2}$\\
COMO-Tanimoto & $98.6$ & $\mathbf{96.5}$ & $\underline{94.8}$ & $\underline{88.4}$ & $\underline{98.0}$ & $\mathbf{93.4}$ & $\mathbf{96.1}$ & $\underline{87.4}$ & $84.6$ & $\underline{77.1}$\\
COMO-Visual & $\underline{98.7}$ & $\underline{96.3}$ & $94.6$ & $87.1$ & $96.8$ & $\underline{93.3}$ & $\underline{96.0}$ & $\mathbf{87.5}$ & $ \underline{86.4}$ & $76.9$ \\    
\bottomrule
\end{tabular}
\end{table}

The upper half of Table~\ref{tab:ablation} presents results for two pure-MLE models. Compared with the baseline trained only on $\mathcal{D}_{\text{MLE}}$, adding $\mathcal{D}_{\text{MRT}}$ under pure MLE changes performance by $-0.9$ to $+1.8$ points across the ten benchmarks, which may fall within plausible run-to-run variation, with slight degradation on half of the ten benchmarks. By contrast, MRT training improves over the same $\mathcal{D}_{\text{MLE}}$-only baseline by up to $+18.1$ points on JPO ($71.0\!\to\!89.1$), $+12.4$ on ACS ($75.8\!\to\!88.2$), and $+10.6$ on WildMol-10K ($66.6\!\to\!77.2$)---margins far beyond such variation---while never falling below the baseline by more than 0.3 points on any benchmark. These results demonstrate that the performance gains of COMO cannot be attributed to the usage of additional real-world data but stem from the MRT training policy.

\subsubsection{Architecture Generalization of MRT}
\label{sec:swinocsr_mrt}

To demonstrate that MRT is architecture-agnostic and not limited to the COMO architecture, we apply the same MRT fine-tuning procedure to a different end-to-end OCSR model, SwinOCSR~\citep{xu2022swinocsr}. For this cross-architecture experiment, we load a publicly released pretrained checkpoint and perform MRT-only fine-tuning for 6 epochs on the same 83K MolParser-SFT dataset used for COMO. The reward configurations are identical to COMO. We use the same hyperparameters as in COMO but lower the sampling temperature to $\tau{=}0.3$ to compensate for SwinOCSR's relatively weaker ability and ensure a more stable output quality. The learning rate is set to $10^{-5}$ for both the encoder and the decoder, with a batch size of 16 and $N{=}32$ samples. 

\begin{table}[h]
\centering
\scriptsize
\caption{SwinOCSR MRT fine-tuning evaluation results.}
\label{tab:swinoccsr}
\begin{tabular}{{l}*{10}{c}}
\toprule
& \multicolumn{2}{c}{Synthetic} & \multicolumn{8}{c}{Real-World} \\
\cmidrule(lr){2-3} \cmidrule(lr){4-11}
& \makecell{Indigo} & \makecell{ChemDraw} & \makecell{CLEF} & \makecell{JPO} & \makecell{UOB} & \makecell{USPTO} & \makecell{USPTO-10K} & \makecell{Staker} & \makecell{ACS} & \makecell{WildMol-10K} \\
\midrule

Pretrained & $67.8$ & $74.5$ & $26.9$ & $12.7$ & $ 49.3$ & $ 24.9$ & $ 32.4$ & $0.2$ & $20.2$ & $10.6$\\
\midrule
Tanimoto & $\mathbf{80.3}$ & $\mathbf{84.4}$ & $\mathbf{73.4}$ & $25.4$ & $\mathbf{98.1}$ & $\mathbf{60.5}$ & $69.8$ & $\mathbf{4.1}$ & $\textbf{40.5}$ & $\mathbf{53.4}$\\
Edit-distance & $79.2$ & $84.2$ & $72.7$ & $\mathbf{27.8}$ & $97.9$ & $58.6$ & $\mathbf{70.1}$ & $4.0$ & $39.0$ & $ 52.5$\\
Visual & $78.4$ & $83.4$ & $ 70.1 $ & $23.4$ & $97.7$ & $57.1$ & $67.1$ & $1.6$ & $37.2$ & $50.2$\\
\bottomrule
\end{tabular}
\end{table}

Table~\ref{tab:swinoccsr} presents the evaluation results before and after MRT fine-tuning. We observe a huge improvement introduced by MRT across benchmarks, increasing exact match accuracy by factors ranging from $2\times$ (from 20.2 to 40.5 on ACS) to $20\times$ on real-world datasets compared with the pretrained SwinOCSR checkpoint. These results showcase that MRT provides substantial gains independent of the underlying OCSR backbone. The ability to consistently improve a pretrained model with minimal changes underscores MRT coupled by our reward design as a broadly applicable, plug-and-play training strategy for enhancing end-to-end OCSR systems. 

\subsubsection{Effect of Joint MLE--MRT schedule}
\label{sec:twostage}

As an alternative to the joint MLE--MRT pipeline, we train and evaluate two-stage variants: 30 epochs of pure MLE followed by 6 epochs of MRT-only fine-tuning, mirroring the transfer protocol of \S~\ref{sec:swinocsr_mrt}. As shown in Table.~\ref{tab:twostage}, interleaving prominently outperforms the two-stage pipeline on real-world benchmarks: the joint model improves over the corresponding two-stage variant by up to $+8.4$\,pp on ACS, with consistent gains on JPO ($+1.6$ to $+4.2$\,pp), WildMol-10K ($+3.3$ to $+3.4$\,pp), CLEF ($+1.1$ to $+2.0$\,pp), and Staker ($+0.8$ to $+1.3$\,pp). The two-stage model is exposed exclusively to $\mathcal{D}_{\text{MRT}}$ during MRT fine-tuning and forfeits the dense token-level supervision that anchors the model to the broader data distribution; the interleaved schedule preserves it by construction.

\begin{table}[h]
\centering
\scriptsize
\caption{Two-stage training versus the joint MLE--MRT pipeline. Top: two-stage exact-match accuracy per reward variant; bottom: the improvement $\Delta$ of the joint pipeline over the corresponding two-stage variant.}
\label{tab:twostage}
\begin{tabular}{{l}*{10}{c}}
\toprule
& \multicolumn{2}{c}{Synthetic} & \multicolumn{8}{c}{Real-World} \\
\cmidrule(lr){2-3} \cmidrule(lr){4-11}
& \makecell{Indigo} & \makecell{ChemDraw} & \makecell{CLEF} & \makecell{JPO} & \makecell{UOB} & \makecell{USPTO} & \makecell{USPTO-10K} & \makecell{Staker} & \makecell{ACS} & \makecell{WildMol-10K} \\
\midrule
\textit{Two-stage} & & & & & & & & & & \\

Edit-Distance & $98.5$ & $96.0$ & $93.0$ & $84.9$ & $98.0$ & $92.9$ & $95.9$ & $86.2$ & $79.8$ & $73.8$ \\
Tanimoto & $98.5$ & $95.8$ & $93.7$ & $86.2$ & $96.9$ & $93.1$ & $96.2$ & $86.6$ & $80.7$ & $73.8$ \\
Visual & $98.5$ & $96.0$ & $93.2$ & $85.5$ & $97.4$ & $93.0$ & $96.2$ & $86.7$ & $80.4$ & $73.5$ \\

\midrule

\textit{$\Delta$ (Joint $-$ Two-stage)} & & & & & & & & & & \\
Edit-Distance & $+0.3$ & $+0.3$ & $+2.0$ & $+4.2$ & $+0.5$ & $+0.1$ & $0.0$ & $+1.3$ & $+8.4$ & $+3.4$\\
Tanimoto & $+0.1$ & $+0.7$ & $+1.1$ & $+2.2$ & $+1.1$ & $+0.3$ & $-0.1$ & $+0.8$ & $+3.9$ & $+3.3$\\
Visual & $+0.2$ & $+0.3$ & $+1.4$ & $+1.6$ & $-0.6$ & $+0.3$ & $-0.2$ & $+0.8$ & $+6.0$ & $+3.4$ \\
\bottomrule
\end{tabular}
\end{table} 

\section{Qualitative Error Analysis}

\begin{figure}[h]
\centering
\includegraphics[width=1\linewidth]{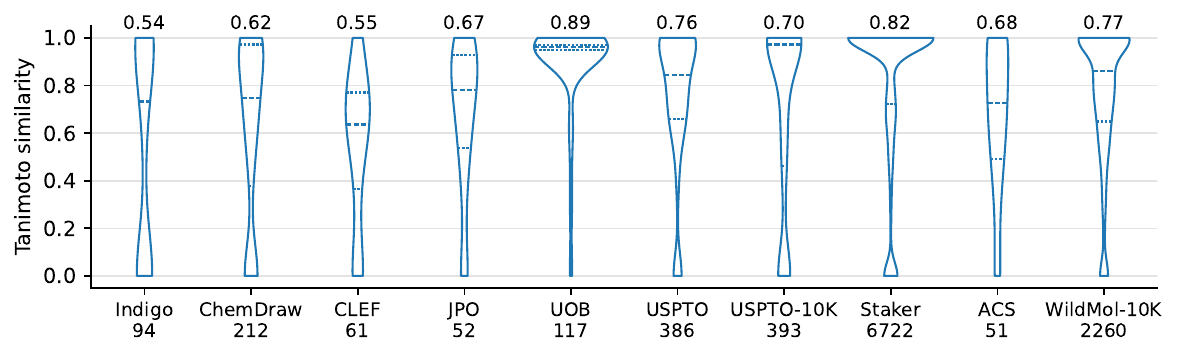}
\caption{Distribution of Tanimoto similarity between wrong predictions and their ground-truth references across all ten benchmarks. Only non–exact-match predictions are included. Lines within each violin mark the lower quartile, median, and upper quartile. The mean Tanimoto similarity is annotated above each violin. The number of incorrect predictions are reported below the benchmark names.}
\label{fig:tanimoto_violin}
\end{figure}

To better understand the remaining failure modes of COMO, we first characterize the overall structural proximity of wrong predictions to their references. Fig.~\ref{fig:tanimoto_violin} shows the distribution of Tanimoto similarity between wrong predictions and their reference SMILES across all ten benchmarks. Across nearly all datasets, the modal Tanimoto similarity for failures is substantially above zero. Distributions on real-world benchmarks such as Staker ($0.82$), USPTO ($0.76$), and WildMol-10K ($0.77$) are heavily left-skewed, indicating that the dominant failure mode is structural near-correctness rather than complete mis-recognition.

We then go further to manually examine non-matching predictions across the evaluation benchmarks. Our analysis reveals two qualitatively distinct categories: \emph{genuine model errors}, where COMO fails to correctly recognize structural features present in the image, and \emph{annotation errors}, where benchmark labels are incorrect and COMO's predictions accurately reflect the visual input. 

\subsection{Genuine Model Prediction Errors}

\begin{figure}[t]
\centering
\includegraphics[width=1\linewidth]{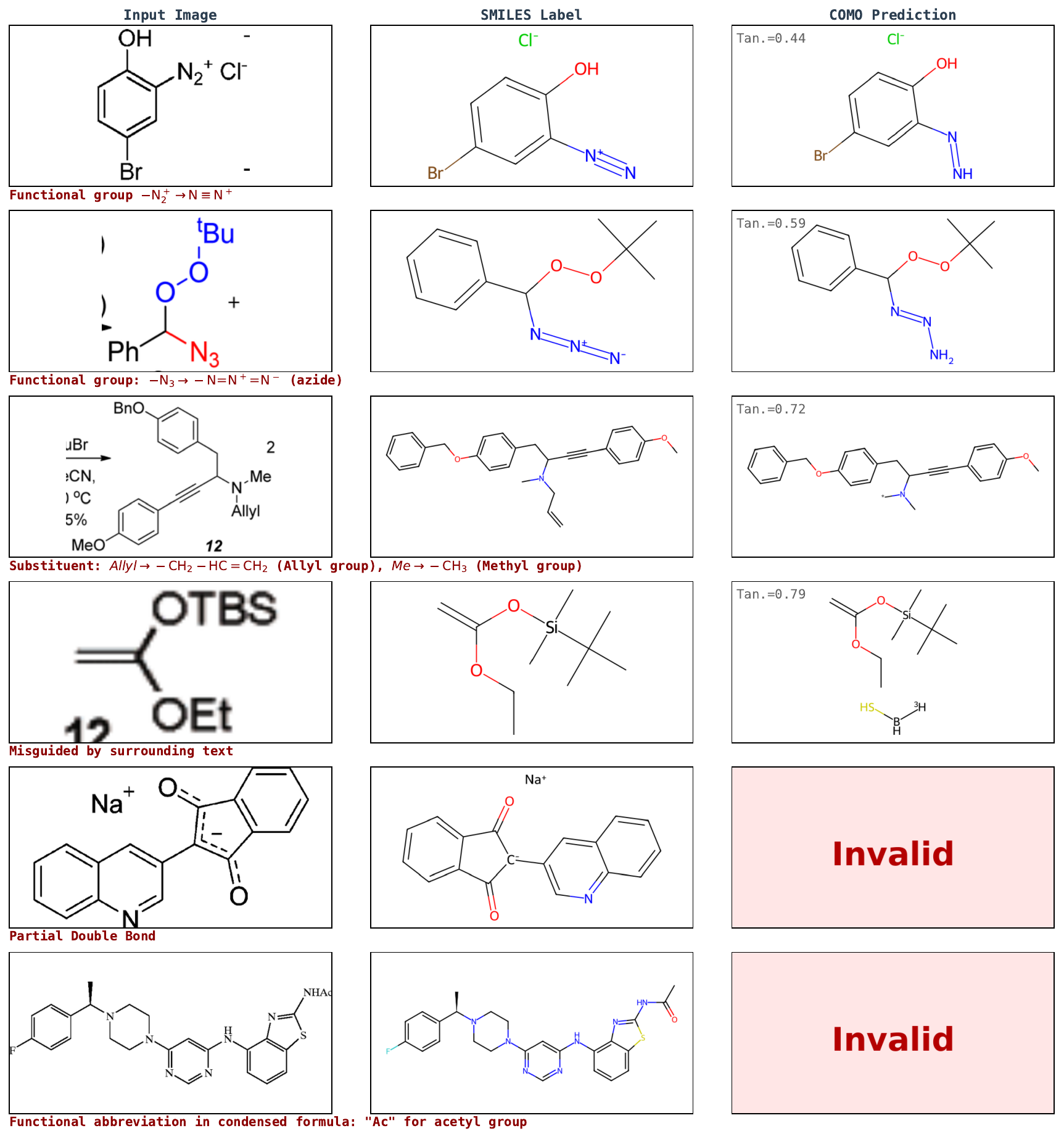}
\caption{Representative failure cases of COMO. Each row shows the input molecular image (left), image rendered from the reference SMILES (center), and image rendered from prediction (right). The Tanimoto similarity to the reference is annotated in each prediction panel.}
\label{fig:error_models}
\end{figure}

Fig.~\ref{fig:error_models} presents six representative failure cases, which fall into the following recurring categories. 

\begin{itemize}
\item \textit{Functional group misidentification} (rows~1--2): COMO confuses two charged nitrogen functional groups with visually similar neutral counterparts. In row~1, the diazonium cation ($\mathrm{N}{\equiv}\mathrm{N}^{+}$) of an aryl diazonium chloride is predicted as an azo linkage (\texttt{N=N}): the model fails to reproduce the triple bond and the formal positive charge on the terminal nitrogen. In row~2, the azide group (\texttt{N=[N+]=[N-]}), expressed in its resonance form with alternating formal charges, is predicted as an uncharged triazene chain (\texttt{N=NN}). Both errors arise because the compact visual signatures of diazonium and azide closely resemble their uncharged counterparts ($\mathrm{N}{=}\mathrm{N}$ and $\mathrm{N}{=}\mathrm{N}{-}\mathrm{N}$, respectively), and because formally charged nitrogen groups are systematically underrepresented in synthetic training data.

\item \textit{Substituent abbreviation failure} (row~3): Real-world chemical diagrams often use shorthand notations that require chemical knowledge to expand. In row~3, the label ``Allyl'', which denotes $-\mathrm{CH}_{2}\mathrm{CH}{=}\mathrm{CH}_{2}$, is not properly resolved by COMO, and the entire substituent is replaced with a wildcard placeholder (\texttt{*}). 

\item \textit{Context contamination} (row~4): COMO correctly identifies the core TBS-protected vinyl ether but appends a spurious fragment, likely decoded from a neighboring textual annotation in the document layout, which prevents an exact match despite the otherwise correct structural recognition.

\item \textit{Uncommon bond and group notations} (rows~5--6): Two cases involve structures that fall outside the model's learned distribution. Row~5 depicts a polycyclic sodium enolate with partial double bonds, an anionic sp$^{3}$ carbon (\texttt{[C-]}), and a dissociated sodium counterion (\texttt{[Na+]}) in a fused indanedione--quinoline scaffold; COMO fails to produce a valid SMILES. Row~6 includes acetamide group ($-\mathrm{C}(=\mathrm{O})\mathrm{CH}_{3}$) abbreviated as ``Ac'' in a condensed formula (NHAc); COMO again produces no valid output. Both cases highlight limitations in handling rare or unseen notational conventions. 
\end{itemize}

\subsection{Annotation errors in benchmark labels}

\begin{figure}[h]
\centering
\includegraphics[width=1\linewidth]{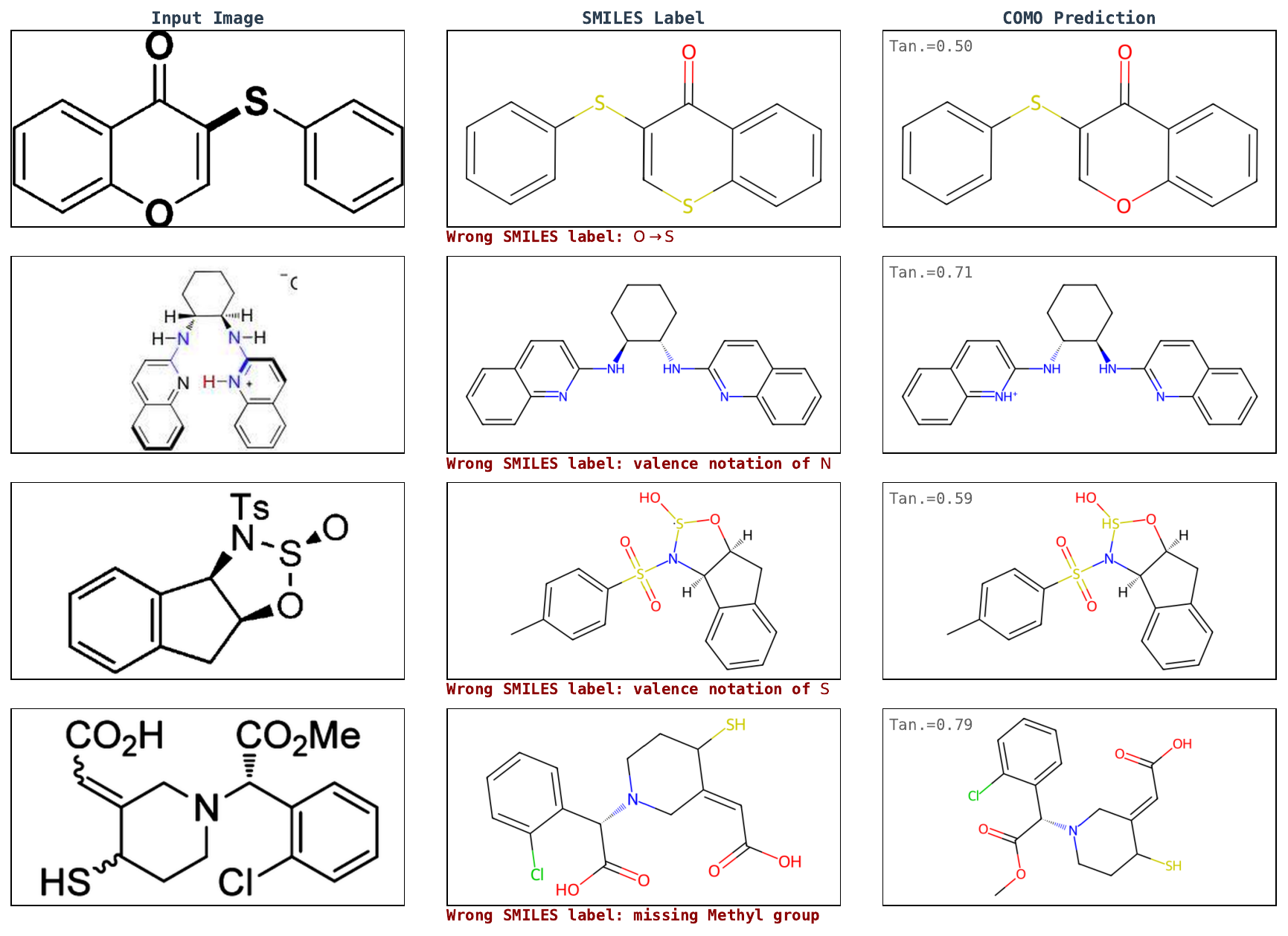}
\caption{Representative annotation errors in OCSR benchmark labels.}
\label{fig:error_labels}
\end{figure}

Fig.~\ref{fig:error_labels} presents four cases where manual inspection indicates that COMO's prediction is structurally correct, but the apparent failure originates from incorrect reference labels. 

Row~1 (ACS benchmark, \#101): the reference SMILES encodes an aromatic sulfur (\texttt{s}) at a specific ring position in a chromone-like bicyclic scaffold, whereas the image clearly shows oxygen (\texttt{o}) at that position; COMO correctly predicts oxygen. 

Row~2 (WildMol-10K, \#2471): the reference SMILES records a neutral quinoline nitrogen, whereas the image depicts a protonated quinolinium nitrogen, consistent with the hydrogen annotated in red; COMO's prediction captures this protonation.

Row~3 (ACS benchmark, op050182n): the cyclic sulfur in a sultam ring is annotated with the bracket atom \texttt{[S]}---a notation that carries zero implicit hydrogen and overrides the default valence, yet with three bond partners in the ring (to nitrogen and two oxygens), the sulfur is left with an unusual trivalent valence. COMO predicts \texttt{[SH]} (adding an additional hydrogen), to fit the common sense.

Row~4 (ACS benchmark, acs.joc.5b00632): the reference encodes a methyl ester terminus as a carboxylic acid (-C(=O)O), which omits the methyl group that is clearly legible in the image as ``CO$_2$Me''; COMO correctly predicts the methyl ester (COC(=O)), receiving a false penalty. 

These cases collectively suggest that near-miss failures, as evidenced by high mean Tanimoto similarities in Fig.~\ref{fig:tanimoto_violin}, are often attributable to label noise rather than genuine model errors. Thus, the true recognition accuracy of COMO on benchmarks such as ACS may be slightly underestimated by exact-match metrics. More broadly, these case analyses underscore the need for systematic label validation in future OCSR benchmark construction. 

\section{Conclusion}

We present COMO, a closed-loop optical molecule recognition framework that integrates Minimum Risk Training (MRT) into an encoder--decoder pipeline. By sampling from the model's own predictions and optimizing a non-differentiable reward, COMO directly targets molecular-level objectives that token-level MLE cannot capture. The reward formulation is modular: we instantiate edit distance, Tanimoto, and visual similarity variants, all of which deliver consistent gains and achieve state-of-the-art results across all 10 benchmarks\footnote{The results of the strongest prior baseline, MolParser, are quoted from its publication since its model is not publicly available, and its WildMol-10K evaluation protocol is undocumented (\S~4.4).}

Two ablation studies support our findings. First, replacing MRT with identically scheduled MLE steps on the same additional data results in performance degradation, confirming that the improvements stem from the closed-loop training mechanism rather than from the additional data. Second, applying MRT to SwinOCSR yields substantial improvements (up to 20$\times$ on real-world benchmarks), demonstrating that MRT is architecture-agnostic and broadly applicable to end-to-end OCSR systems. 

Despite these advances, several limitations suggest directions for future work. First, the visual reward variant is best viewed as a demonstration that the reward interface accepts learned perceptual signals: its accuracy is on par with, but not superior to, the string-based rewards across all ten benchmarks, and a post-hoc subpopulation analysis (stereochemistry, fused polycyclics, the Staker subset) reveals no stratum with a distinct advantage. A plausible explanation---consistent with the hypothesis that rendered images preserve structural details compressed away by string or fingerprint metrics---is the limited representation power of the MolDiscriminator; more expressive visual encoders could further unlock the potential of perceptual rewards. Second, the current framework remains dependent on labeled data, which may itself contain errors, as has already been observed in existing datasets. These issues limit large-scale training and reduce its effectiveness by introducing noisy or conflicting supervision. Future work should explore weakly supervised or unsupervised approaches to scale the training corpus while mitigating the impact of annotation errors. Third, our image augmentation strategy could be enriched with targeted visual contaminants~\citep{chen2024molnextr}, such as reaction arrows, to better simulate the complex layouts and artifacts found in real-world documents. Fourth, due to computational constraints, we use a relatively small MRT sample size ($N{=}32$), below the recommended value in \cite{shen2016minimum}. Future work is needed to explore larger $N$ or more sophisticated sampling strategies, such as best-$N$ beam search or adaptive sampling, to further enhance the efficiency and effectiveness of closed-loop optimization. Finally, each configuration is trained once, and quantifying seed-level variance is left to future work.

\bibliography{main}

\end{document}